# Robot gains Social Intelligence through Multimodal Deep Reinforcement Learning

Ahmed Hussain Qureshi, Yutaka Nakamura, Yuichiro Yoshikawa and Hiroshi Ishiguro

*Abstract*— For robots to coexist with humans in a social world like ours, it is crucial that they possess human-like social interaction skills. Programming a robot to possess such skills is a challenging task. In this paper, we propose a Multimodal Deep Q-Network (MDQN) to enable a robot to learn human-like interaction skills through a trial and error method. This paper aims to develop a robot that gathers data during its interaction with a human, and learns human interaction behavior from the high dimensional sensory information using end-to-end reinforcement learning. This paper demonstrates that the robot was able to learn basic interaction skills successfully, after 14 days of interacting with people.

## I. INTRODUCTION

Human-robot interaction (HRI) is an emerging field of research with the aim to integrate robots into human social environments. One of the biggest challenges in the development of social robots is to understand human social norms [1]. It is therefore essential for social robots to possess deep models of social cognition, and be able to learn and adapt in accordance with their shared experiences with human partners. Most of the social robots to date are either pre-programmed, or are controlled by teleoperation or semi-autonomous teleoperation [2], and do not possess the ability to learn and update themselves.

Designing an adaptable and autonomous sociable robot is particularly challenging, as the robot needs to correctly interpret human behaviors as well as respond appropriately to them. This is necessary to ensure safe, natural and effective human-robot interaction. Arguably, most of the so-called social robots have limited social interaction skills. One of the main reasons for this limited capability is the diversity in human behavior [3]. Social interaction between humans relies on intention inference such as inferring the intention from walking trajectories, direction of eye gaze, facial expressions, body language and activity in progress. Programming a robot to recognize human intention from the aforementioned factors and respond to diverse human behaviors is notoriously difficult, as it is hard to envision each and every one of the countless possible interaction scenarios. Therefore, it is necessary for a social robot to possess a self-learning paradigm [4] which enables it to learn deep

* This work is partly supported by JSPS Grant-in-Aid for Young Scientists (B) 14444719.

A. H. Qureshi, Y. Nakamura, Y. Yoshikawa and H. Ishiguro are with Department of System Innovation, Graduate School of Engineering Science, Osaka University, 1-3 Machikaneyama, Toyonaka, Osaka, Japan. {qureshi.ahmed, nakamura, yoshikawa, ishiguro}@irl.sys.es.osaka-u.ac.jp

A. H. Qureshi, Y. Yoshikawa and H. Ishiguro are also with JST ERATO ISHIGURO Symbiotic Human-Robot Interaction Project.

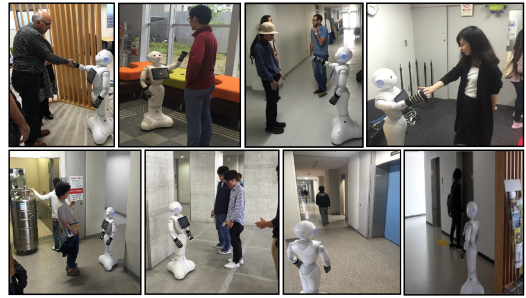

Fig. 1: Robot learning social interaction skills from people.

models of human social cognition from features extracted automatically from high-dimensional sensory information.

Recently, the field of deep learning, also known as representation learning, has emerged and it has achieved many breakthroughs on various tasks of computer vision [5] [6] [7] and speech recognition [8] [9]. Deep learning methods take raw sensory information as input and process it to learn multiple levels of representation automatically, where each level of representation corresponds to a slightly higher level of abstraction [5] [6]. Further advancements in machine learning have merged deep learning with reinforcement learning (RL) which has led to the development of the deep Q-network (DQN) [10] . DQN utilizes an automatic feature extractor called deep convolutional neural network (Convnets) to approximate the action-value function of Q-learning method [10]. DQN has demonstrated its ability to learn from high-dimensional visual input to play arcade video games at human and superhuman level. However, the applicability of DQN to real world human-robot interaction problems has not been explored yet. In this research, we augment our robot with a multimodal deep Q-network (MDQN) which enables the robot to learn social interaction skills through interaction with humans in public places.

The proposed MDQN uses two streams of convolutional neural networks for action-value function estimation. The dual stream convnets process the depth and grayscale images independently. We consider a scenario in which the robot learns to greet people using a set of four legal actions, i.e., waiting, looking towards human, waving hand and handshaking. The objective of the robot is to learn which action to perform in each situation. We conducted the experiment at different locations such as a cafeteria, department reception, various common rooms, etc., as shown in figure 1. After 14 days of interacting with people, the robot exhibited a remarkable level of basic social intelligence. The robot social interaction skills were also evaluated on test data not seen by the system during training.

With this paper we release the source code and the depth dataset[1] collected during the experiment. During the experiment, we collected both grayscale and depth frames but due to privacy concerns we are only releasing the depth dataset.

The rest of the paper is organized as follows. Section II provides discussion on the related work, section III provides a brief background of reinforcement learning and the DQN architecture, section IV explains the proposed multimodal deep Q-learner architecture while section V describes the experimental setup. Section VI and VII give the results of the experiment and the discussion on what the robot has learned through interaction with humans, respectively. Finally, section VIII concludes the paper and also suggests some future areas of research in this particular domain.

## II. RELATED WORK

The proposed work utilizes deep Q-learning to enable a robot to learn social interaction skills from experience interacting with people. This section describes related research from the fields of human-robot interaction, deep learning and deep reinforcement learning.

The most relevant prior work in HRI includes the work by Amor et al. [11] [12], Lee et al. [13], and Wang et al. [14]. The proposed methods in [11] [12] [13] learn responsive robot behavior by imitating human interaction partners. The movements of two persons, action-reaction pairs, are recorded during the human-human interaction with a motion capture system. An interaction model is learned from the data, which enables a robot to compute the best response to a human interaction partner's current behavior. However, the motion capture system used for data recording is not user-friendly and it does not yield natural human behavior, as the participants have to wear track-able markers. In [14], the authors present a probabilistic graphical model with intentions represented as latent states, where the mapping from observations to latent states is approximated by a Gaussian Process. The proposed model allows intention inference from observed movements. However, we believe that, in the case of HRI, intention inference is not dependent on body movements only but also on eye gaze, body language, walking trajectories, activity in progress etc. Therefore, limiting intention inference to body movements alone does not seem promising. Furthermore, the prior art stated above considers only one human interaction partner for the robot in the scene. In this paper, more complex scenarios are considered where the robot can be approached by a group of people willing or not willing to interact with it.

So far, the deep learning and deep reinforcement learning (DRL) research has been applied to areas, though include robotics, which have little to do with the domain of human-robot interaction. Also, most of these applications are limited to simulated environments. Predicting human intention from video data has only recently been addressed in deep learning literature [15]. Our work differs from aforementioned work because, in our case, the robot is acting in a real, uncontrolled environment; by taking an action the robot may affect the human intention. Therefore, the robot has to perceive human behavior, as well as, its own actions according to human social norms. From the domain of DRL, the idea of deep Q-learning has recently been extended to the robotics field to solve twenty continuous control problems; such as legged locomotion, car driving, cartpole swing-up etc.[16]. However, these methods have not yet been extended to the domain of human-robot interaction or in real world environments. To the best knowledge of the authors there does not exist any work that utilizes deep learning coupled with reinforcement learning for realization of physical human-robot social interaction.

## III. BACKGROUND

We consider a standard reinforcement learning formulation in which an agent interacts sequentially with an environment $E$ with an aim of maximizing cumulative reward. At each time-step, the agent observes a state $s_t$, takes an action $a_t$ from the set of legal actions $\mathcal{A} = \{1, \cdots, K\}$ and receives a scalar reward $r_t$ from the environment.

An agent's behavior is formalized by a policy $\pi$, which maps states to actions. The goal of a RL agent is to learn a policy $\pi$ that maximizes the expected total return. The expected total return is the sum of rewards discounted by factor $\gamma : [0, 1]$ at each time-step ($\gamma = 0.99$ for the proposed work) i.e., $R_t = \sum_{t'=t}^{T} \gamma^{t'-t} r_{t'}$, where $T$ is the step at which the agent's interaction with the environment terminates. Furthermore, the *action-value function* $Q^\pi(s, a)$ is the expected return when taking the action $a$ in state $s$ under the policy $\pi$, $Q^\pi(s, a) = \mathbb{E}[R_t | s_t = s, a_t = a, \pi]$. The maximum expected return that can be achieved by following any policy is given by the optimal action-value function $Q^*(s, a) = \max Q^\pi(s, a)$. The optimal action-value function obeys a fundamental recursive relationship known as the *Bellman equation:* $Q^*(s, a) = \mathbb{E}[r + \gamma \max_{a'} Q^*(s', a') | s, a]$. The intuition behind it is that: given that the optimal action-value function $Q^*(s', a')$ of the sequence $s'$ at next time-step is deterministic for all possible actions $a'$, the optimal policy is to select an action $a'$ that maximizes the expected value of $r + \gamma Q^*(s', a')$.

One of the practices in RL, especially Q-learning, is to estimate the action-value function by using a function estimator such as neural networks i.e., $Q(s, a) \approx Q(s, a, \theta)$. The parameters $\theta$ of the neural Q-network are adjusted iteratively towards the Bellman target. Recently, a new approach to approximate action-value function, called deep Q-networks (DQN), has been introduced which is much more stable than previous techniques. In DQN, the action-value function is approximated by a deep convolutional neural network. The DQN technique for function approximation differs from previous methods in two ways: 1) It uses experience replay [17] i.e., it stores the agent's interaction experience, $e_t = (s_t, a_t, r_t, s_{t+1})$, with the environment into the replay memory, $\mathcal{M} = e_1, \cdots, e_t$, at each time-step; 2) It maintains two Q-networks: the Bellman target is computed by the target network with old parameters i.e., $Q(s, a; \theta^-)$, while the learning network $Q(s, a; \theta)$ keeps the current parameters which may get updated several times at each time-step. The old parameters $\theta^-$ are updated to current parameters $\theta$ after every C iterations.

---

[1] To get the source code and the dataset please visit https://sites.google.com/a/irl.sys.es.osaka-u.ac.jp/member/home/ahmed-qureshi/deephri

In DQN, the parameters of the Q-network are adjusted iteratively towards the Bellman target by minimizing the following loss function:

$$L_i(\theta_i) = \mathbb{E}\left[\left(r + \gamma \max_a' Q(s', a'; \theta_i^-) - Q(s, a; \theta_i)\right)^2\right] \quad (1)$$

For each update, $i$, a mini-batch is sampled from the replay memory. The current parameters $\theta$ are updated by stochastic gradient descent in the direction of the gradient of the loss function with respect to the parameters i.e.,

$$\nabla L_i(\theta_i) = \mathbb{E}\left[\left(r + \gamma \max_a' Q(s', a'; \theta_i^-) - Q(s, a; \theta_i)\right) \nabla_{\theta_i} Q(s, a; \theta)\right] \quad (2)$$

Finally, the agent's behavior at each time-step is selected by an $\epsilon$-greedy policy where the greedy strategy is adopted with probability $1 - \epsilon$ while the random strategy with probability $\epsilon$.

## IV. THE PROPOSED ALGORITHM

The proposed algorithm consists of two streams that work independently: one for processing the grayscale frames, and another for the depth frames. Algorithm 1 outlines the proposed method. Since the model is dual stream, therefore, the parameters $\theta$ and $\theta^-$ consist of parameters of both networks. Unlike DQN [10], we separate the data generation and training phase. Each day of experiment corresponds to an episode during which the algorithm executes both the data generation phase and the training phase. Following is a brief description of both phases.

*Data generation phase:* During the data generation phase, the system interacts with the environment using Q-network $Q(s, a; \theta)$. The system observes the current scene, which comprises of grayscale and depth frames, and takes an action using the $\epsilon$-greedy strategy. The environment in return provides the scalar reward (please refer to section 5(2) for the definition of reward function). The interaction experience $e = (s_i, a_i, r_i, s_{i+1})$ is stored in the replay memory $\mathcal{M}$. The replay memory $\mathcal{M}$ keeps the $N$ most recent experiences which are later used by the training phase for updating the network parameters.

*Training phase:* During the training phase, the system utilizes the collected data, stored in replay memory $\mathcal{M}$, for training the networks. The hyperparameter $n$ denotes the number of experience replay. For each experience replay, a mini buffer $\mathcal{B}$ of size 2000 interaction experiences is randomly sampled from the finite sized replay memory $\mathcal{M}$. The model is trained on the mini batches sampled from buffer $\mathcal{B}$ and the network parameters are updated iteratively in the direction of the bellman targets. The random sampling from the replay memory breaks the correlation among the samples since the standard reinforcement learning methods assume the samples are independently and identically distributed. The reason for dividing the algorithm into two

---

**Algorithm 1:** Mutlimodal Deep Q-learner.

1 Initialize replay memory $\mathcal{M}$ to size $N$
2 Initialize training Q-network $Q(s, a; \theta)$ with parameters $\theta$
3 Initialize target Q-network $\hat{Q}(s, a; \theta^-)$ with weights $\theta^- = \theta$
4 **for** episode $= 1, M$ **do**
5    **Data generation phase:**
6    Initialize the start state to $s_1$
7    **for** $i = 1, T$ **do**
8       With probability $\epsilon$ select a random action $a_t$ otherwise select $a_t = \max_a Q(s_t, a; \theta)$
9       $s_{t+1}, r_t \leftarrow \text{ExecuteAction}(a_t)$
10      Store the transition $(s_t, a_t, r_t, s_{t+1})$ in $\mathcal{M}$
11    **Training phase:**
12    Randomize a memory $\mathcal{M}$ for experience replay
13    **for** $i = 1, n$ **do**
14       Sample random minibuffer $\mathcal{B}$ from $\mathcal{M}$
15       **while** $\mathcal{B}$ **do**
16          Sample minibatch $m$ of transitions $(s_k, a_k, r_k, s_{k+1})$ from $\mathcal{B}$ without replacement

$$y_k = \begin{cases} r_k, \text{if step k+1 is terminal} \\ r_k + \gamma \max_a \hat{Q}(s_{k+1}, a; \theta^-), \text{otherwise} \end{cases}$$

              Perform gradient descent on loss $(y_k - Q(s_k, a_k; \theta))^2$ w.r.t the network parameters $\theta$
17    After every C-episodes sync $\theta^-$ with $\theta$.

---

phases is to avoid the delay that would be caused if the network were trained during the interaction period. The DQN agent [16] works in a cycle in which it first interacts with the environment and stores the transition into the replay memory, then it samples the mini batch from the replay memory and trains the network on this mini batch. This cycle is repeated until termination occurs. The sequential process of interaction and training can be acceptable only in fields other than HRI. In HRI, the agent has to interact with people based on social norms, so, any pause or delay while the robot is on the field is unacceptable. Therefore, we divide the algorithm into two stages: in the first stage, the robot gathers data through interaction for some finite period of time, in the second stage, it goes to its rest position. During the resting period, the training phase gets activated to train the multimodal deep Q-network (MDQN).

## V. IMPLEMENTATION DETAILS

This section formally describes implementation details of the research. The MDQN agent was implemented in torch/lua[2], while robot actions were implemented using python. The entire experiment was performed using 3.40GHz×8 Intel Core i7 processor with 32 GB ram and GeForce GTX 980 graphic processing unit. The rest of the section explains the robotic system, MDQN model architecture, visual information pre-processing details, experimental details and evaluation procedure.

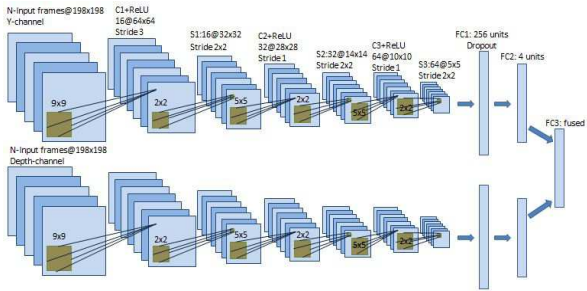

Fig. 2: Dual stream convolutional neural network

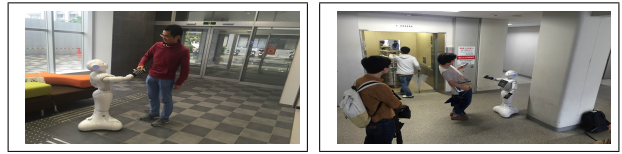

(a) Successful handshake.  (b) Unsuccessful handshake.

Fig. 3: Instances of successful and unsuccesful handshakes.

### A. Robotic system

Aldebaran's Pepper robot [3] was used for the proposed research. Pepper has two built-in 2D cameras and one 3D sensor. Although Pepper has two 2D cameras, only the top camera, located on Pepper's forehead, was used in this research. The ASUS Xtion 3D sensor situated behind the robot eyes was utilized for depth images. Both the top 2D camera and the 3D sensor returned images with resolution $320 \times 240$ at 10 frames per second. Moreover, the robot's right hand was equipped with an external FSR touch sensor which was hidden under soft woolen gloves for aesthetic reasons. In addition, the robot was augmented with 1) four set of actions through which it can interact with the people; 2) a reward function with which the robot can evaluate how well it is performing. Following subsections formally describe the robot actions and the reward function.

*1) Robot actions:* This paragraph provides the actions definition and their implementation details. The action set comprised of four legal actions, i.e., waiting, looking towards humans, waving its hand and hand shaking with a human. The description of the actions is as follows:

*Wait:* For waiting, the robot randomly picks the head orientation from the allowable range of head pitch and head yaw. During this action, no attempt to engage the human into the interaction is made.

*Look towards human:* This action makes the robot sensitive to the stimuli coming from the environment. If robot senses any stimulus, it looks at the stimulus origin and checks if there is any human there or not. In the case of human presence, the robot tracks the person with its head in order to engage him/her for the interaction otherwise, the robot returns to its previous orientation. The stimuli used to instill awareness into the robot are the sound detection and the movement detection.

[2]http://torch.ch/
[3]//www.aldebaran.com/en/cool-robots/pepper/find-out-more-about-pepper

*Wave hand:* This is a simple hand waving gesture. During its execution, the robot says *Hello* or *Hi*, and attempts to gain peoples' attention by tracking them with its head.

*Handshake:* In handshaking action, the robot raises its hand to a certain height and waits at this position for a few seconds. If the external touch sensor, on the robot's hand, signals the touch, then the robot grabs the person's hand and says *Nice to meet you*, otherwise, the robot's hand goes back to its previous position. Moreover, while performing this action, the robot adjusts its body rotation and head position in order to track the target person from whom it may get the handshake.

*2) Reward function:* The external touch sensor on the robot's right hand detects if a handshake has happened or not. This forms the basis for the reward function. The robot gets a reward of 1 on the successful handshake, -0.1 on an unsuccessful handshake and 0 for the rest of the three actions. Figures 3(a) and 3(b) depict example scenarios of successful and unsuccessful handshakes respectively. In the scenario shown in figure 3(a), the handshake happens successfully therefore the agent gets the reward value 1, whereas in the situation shown in figure 3(b), the person is taking the robot's picture while the robot is attempting to shake their hand; since this is an in appropriate social reaction, the agent will be rewarded with -0.1.

### B. Model Architecture

The proposed model comprises of two streams, one for the gray-scale information, and another for the depth information. The structure of the two streams is identical and each stream comprises of eight layers (excluding the input layer). The overall model architecture is schematically shown in figure 2. The inputs to the y-channel and the depth channel of the multimodal Q-network are grayscale ($198 \times 198 \times 8$) and depth images ($198 \times 198 \times 8$), respectively. Since each stream takes eight frames as an input, therefore, the last eight frames from the corresponding camera are pre-processed and stacked together to form the input for each stream of the network. Since the two streams are identical so we only discuss the structure of one of the streams. The $198 \times 198 \times 8$ input images are given to first convolutional layer (C1) which convolves 16 filters of $9 \times 9$ with stride 3, followed by rectifier linear unit function (ReLU) and results into 16 feature maps each of size $64 \times 64$ (we denote this by $16@64 \times 64$). The output from C1 is fed into sub-sampling layer S1 which applies $2 \times 2$ max-pooling with the stride of $2 \times 2$. The second (C2) and third (C3) convolutional layer convolve 32 and 64 filters, respectively, of size $5 \times 5$ with stride 1. The output from C2 and C3 passes through the non-linear ReLU function and is fed into sub-sampling layers S2 and S3, respectively. The final hidden layer is fully connected with 256 rectifier units. The output layer is fully-connected linear layer with 4 units, one unit for each legal action.

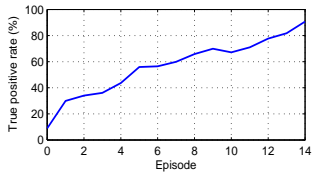

Fig. 4: MDQN performance on test dataset over the series of episodes.

| Trained Model | MDQN | y-channel | depth-channel |
|---|---|---|---|
| Accuracy (%) | 95.3 | 85.9 | 82.6 |
| True positive rate (%) | 90.7 | 71.8 | 66.2 |
| False positive rate (%) | 3.09 | 9.4 | 11.3 |
| Misclassification rate (%) | 4.6 | 14.9 | 16.9 |

TABLE I: Performance measures of trained Q-networks.

### C. Pre-processing

The pre-process function prepares the input appropriately for the model architecture. The robotic system provides the grayscale and the depth images of size $320 \times 240$ at the frame rate of 10 fps. The pre-process function rescales the grayscale and the depth frame to $198 \times 198$. This pre-processing is executed on the eight most recent grayscale and depth frames, which are then stacked together to form the input for each stream of the dual stream Q-network.

### D. Experiment details

The proposed method is divided into two phases, i.e., the data generation phase and the training phase. For every episode, the algorithm passes through these two phases.

During the data generation phase, the robot interacts with the environment for around 4 hours (we call it the interaction period). During the interaction period, the number of steps $i$ (see Algorithm 1) executed by the robot depended on the internet speed[4] since the communication between Pepper and the computer system on which the MDQN was implemented occured over the wireless internet. The behavior strategy during this phase is $\epsilon$-greedy, where $\epsilon$ anneals linearly from 1 to 0.1 over 28,000 steps and then remains at 0.1 for the rest of the steps. For taking the greedy action the outputs from each stream of the dual stream Q-network were fused together and the action with the highest Q-value was selected. For the fusion of outputs from each stream of the Q-network, the algorithm first normalizes the Q-values from each stream and then takes an average of these normalized Q-values. After the interaction period is over, the robot goes to sleep and the training phase begins.

The training procedure presented here is the variant of [10]. The network parameters are trained on mini batches $m$, each of size 25 samples, using the RMSProp algorithm. It should be noted that both network streams, grayscale and depth, were trained independently without any fusion of Q-values, however, the Q-values from each stream were fused during the data generation phase for taking the greedy action. In this presented work, the model was trained over 111,504 grayscale and depth frames, and for each episode, the algorithm performed ten experience replays i.e., $n = 10$. The parameters of target Q-network $\theta^-$ were updated after every episode i.e., $C = 1$.

### E. Evaluation

For testing the model performance, a separate test dataset, comprising 4480 grayscale and depth frames not seen by the system during learning was collected. Since, for every scenario there can be more than one action that can be chosen with utmost propriety, therefore, the agent's decision was evaluated by three volunteers. A sequence of eight frames depicting the scenario and the agent's decision were shown to the volunteers. Each volunteer was asked to judge if the agent's decision was right or not. If the agent's decision was considered wrong by the majority then the evaluators were asked to consent on the most appropriate action for that particular scenario.

## VI. RESULTS

This section summarizes the results of the trained Q-network (agent) on the test dataset. We evaluated the trained y-channel Q-network, depth-channel Q-network and the MDQN on the test dataset; table 1 summarizes the performance measures of these trained Q-networks. In table 1, accuracy corresponds to how often the predictions by the Q-networks were correct. The true positive rate corresponds to the percentage of predicting positive targets as positive and the false positive rate is the percentage of negative instances that were classified as positive. Misclassification rate denotes how often network predictions were wrong. In table 1, it can be seen that the multimodal deep Q-network (Fused) achieved maximum accuracy of 95.3 %, whereas the y-channel and the depth-channel of Q-networks achieved 85.9% and 82.6% accuracy, respectively. Hence, the results in table 1 validate that fusion of two streams improves the social cognitive ability of the agent. Figure 4 shows the performance of MDQN on the test dataset over the series of episodes. The episode 0 on figure 4 corresponds to the Q-network with randomly initialized parameters. The plot indicates that the performance of MQDN agent on test dataset is continuously improving as the agent gets more and more interaction experience with humans. Rest of the section provides the visual evidences of the proposition that the robot gained human-like social intelligence through interaction with humans.

In figures 5-7, the actions wait, look towards human, wave hand, and shake-hand are denoted as W, L, H, and S respectively. For figures 5 and 7, each sub-figure shows

---
[4] With approximately ↓37/↑23 Mbps internet speed the robot could gather $i = 2010$ interaction experiences $e = (s_i, a_i, r_i, s_{i+1})$ in 4 hours. During 14 days of the interaction period, the robot executed 13938 steps in total.

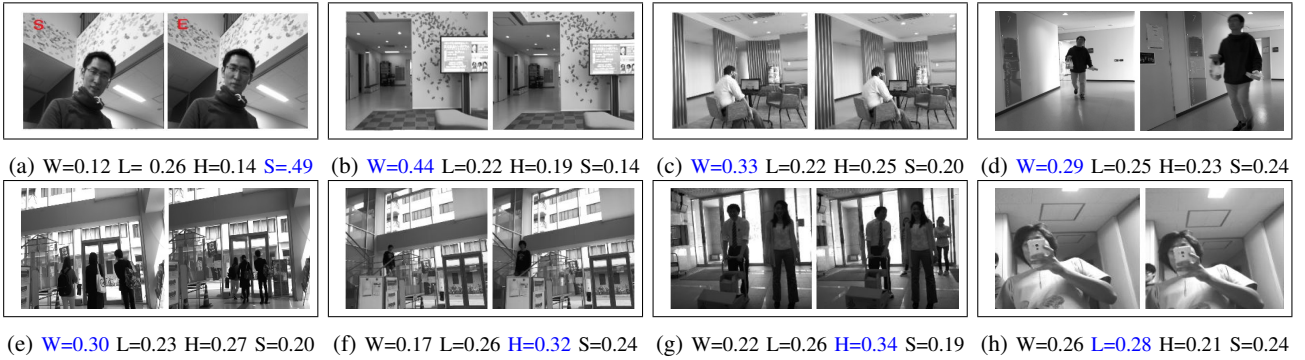

(a) W=0.12 L=0.26 H=0.14 S=.49  (b) W=0.44 L=0.22 H=0.19 S=0.14  (c) W=0.33 L=0.22 H=0.25 S=0.20  (d) W=0.29 L=0.25 H=0.23 S=0.24

(e) W=0.30 L=0.23 H=0.27 S=0.20  (f) W=0.17 L=0.26 H=0.32 S=0.24  (g) W=0.22 L=0.26 H=0.34 S=0.19  (h) W=0.26 L=0.28 H=0.21 S=0.24

Fig. 5: Successful cases of agents decision.

the start (S) and the end (E) frame out of the total eight most recent frames for any situation.

Figures 5 and 6 indicate the instances of successful predictions by the MQDN based agent. The action highlighted in blue shows the action with maximum Q-value, hence indicates the agent's decision for that particular scenario.

In figure 5(a), the person is standing right in front of the robot, therefore, the agent chooses the handshake action. For scenarios depicted in figures 5(b)-5(e) the agent decides to wait. This is because, in the scenario shown in figure 5(b), there is no human in the scene; in case of figure 5(c), the person is working on their laptop; in case of figure 5(d), the person is carrying some things and their hands are not free; and in case of figure 5(e), the group of people are walking away from the robot. Figures 5(f) and 5(g) represent the situation in which the agent chooses the wave-hand action, and looking towards the human action, respectively. Finally, figure 5(h) shows the situation in which the person is standing in front of the robot, but taking the robot's picture therefore the agent decided to look towards him instead of shaking-hand.

Figure 6 shows the events (A-E) that happened sequentially. For each event, only the last frame is presented. In the event A, there is no human for the interaction hence the agent decides to wait. In event B, two people appeared in the scene and the agent switched to the looking towards human action. Following event B, to further get the humans attention, the agent chose the wave-hand action in event C. Event D indicates that the agent has successfully gained the attention of the human as it led to the successful handshake. Finally, in event E, the person's head orientation is not towards the robot so the agent chooses the look towards human action in order to gain their attention again. Figure 7 represents some of the wrong decisions taken by the agent. The action highlighted in red indicates the agent decision while the action highlighted in green represent the decision considered appropriate by the evaluators.

## VII. Discussion

This section provides brief discussion on i) some of the exciting features that the agent (Q-network) has learned through the experiment; and ii) the effect of the reward function on the robot's behavior. Sections A-C highlight that the agent has gained understanding of some of the factors that form the basis for intention inference such as activity in progress, walking trajectories and head orientations. Section E provides a discussion on the effect of the reward function on the robot's social interaction skills.

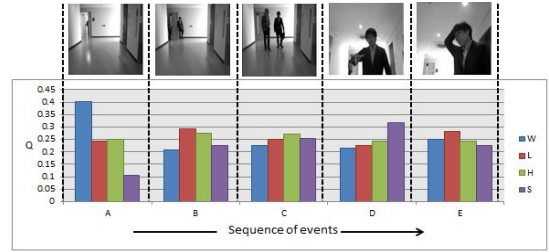

Fig. 6: Series of events(A to E) happened in a sequence.

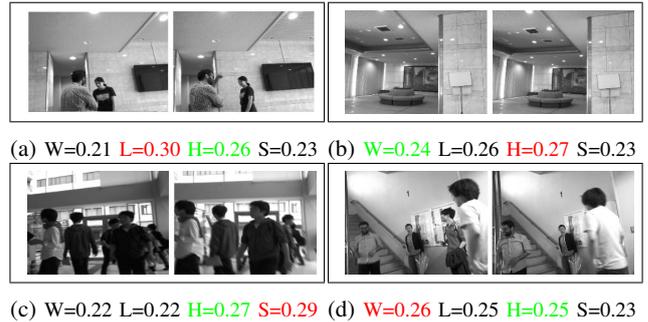

(a) W=0.21 L=0.30 H=0.26 S=0.23  (b) W=0.24 L=0.26 H=0.27 S=0.23

(c) W=0.22 L=0.22 H=0.27 S=0.29  (d) W=0.26 L=0.25 H=0.25 S=0.23

Fig. 7: Unsuccessful cases of agents decision.

### A. Activity in progress

The scenarios shown in figures 5(c), 5(d) and 5(h) show a person working on a laptop, a person carrying some things and a person taking a picture respectively. The agent's decision, during these activities, indicates that it has learned to recognize the activity in progress and has also learned that any interaction during these activities would not lead to the successful handshake, hence agent decides to wait.

### B. Walking trajectory

The agent's decision in situations shown in figures 5(e) and 5(f) shows that it has gained insight about the walking trajectories. In the figure 5(e) people walked away from the robot and in the figure 5(f) a person is coming downstairs and is getting closer to the robot. In the former, MDQN-agent decides to wait as it is quite less probable to get the handshake in that situation, whereas in the latter it decides to wave-hand as there is a chance to get the handshake by gaining the attention of the oncoming person.

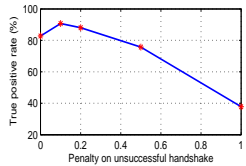

Fig. 8: Effect of reward function on the robot's behavior.

*C. Head orientation*

In figure 6, event D and event E show two different scenarios; one in which the person's head orientation is towards the robot; and other in which it is not towards the robot. For event D, the agent decides to shake-hand while for event E it decides to gain human attention by looking towards them. Hence, this gives an indication that the agent has also learned implications of head orientation on social human-robot interaction.

*D. Effect of reward function on the robot's behavior*

All the results presented so far are based on the reward function discussed earlier. This section formalizes the effect of the reward function on the agent's behavior. Varying the penalty on unsuccessful handshake from 0 to -1 changes the robot behavior from amiable to rude as when the penalty is 0 the robot always tries to handshake and when it is -1, the robot is reluctant to handshake. To understand which behavior is acceptable by humans, we trained five networks with five different reward functions and these five networks were evaluated following the evaluation procedure mentioned earlier. For each reward function the agent gets 0 reward on actions other than handshake, +1 on successful handshake and 0,-0.1,-0.2,-0.5 or -1 on unsuccessful handshake. Figure 8 represents the plot of the true positive rate of each model on test dataset versus corresponding penalty given on unsuccessful handshake. The result shows that the reward function with -0.1 penalty achieved maximum accuracy on the test dataset.

## VIII. CONCLUSION

In social physical human-robot interaction, it is very difficult to envisage all the possible interaction scenarios which the robot can face in the real-world, hence programming a social robot is notoriously hard. To tackle this challenge, we presented a multi-model deep Q-network (MDQN) with which the robot learns the social interaction skills through trial and error method. The results show the diversity of interaction scenarios, which were definitely hard to imagine, and yet the robot was able to learn which action to choose at each time-step in these diverse scenarios. Furthermore, the results also insinuate that the MDQN-agent has learned to give importance to walking trajectories, head orientation, body language and the activity in progress in order to decide its best action.

In our future work, we plan to i) increase the action space instead of limiting it to just four actions; ii) use recurrent attention model so that the robot can indicate its attention; iii) evaluate the influence of three actions, other than handshake, on the human behavior.